\documentclass[11pt,a4paper]{article}
\usepackage{times,latexsym}
\usepackage{url}
\usepackage[T1]{fontenc}
\usepackage{colortbl}
\usepackage{graphicx}
\usepackage{amsmath}
\usepackage{amsfonts}
\usepackage{booktabs}
\usepackage{multirow}
\usepackage{multicol}
\usepackage[export]{adjustbox}
\DeclareUnicodeCharacter{2212}{-}
\usepackage[acceptedWithA]{tacl2021v1}
\definecolor{mygreen}{RGB}{59,205,62}
\definecolor{myred}{RGB}{239,0,0}

\newcommand*{\BigImage}[1]{\includegraphics[width=0.58cm,height=!,valign=m]{#1}}
\newcommand*{\SmallImage}[1]{\includegraphics[width=0.32cm,height=!,valign=m]{#1}}
\newcommand*{\TinyImage}[1]{\includegraphics[width=0.28cm,height=!,valign=m]{#1}}
\usepackage{xspace,mfirstuc,tabulary}

\newif\iftaclinstructions
\taclinstructionsfalse 
\iftaclinstructions
\renewcommand{\confidential}{}
\renewcommand{\anonsubtext}{(No author info supplied here, for consistency with
TACL-submission anonymization requirements)}
\newcommand{\instr}
\fi

\iftaclpubformat 

\else

\fi

\makeatletter
\g@addto@macro\normalsize{%
  \abovedisplayskip 1pt plus 1pt minus 1pt%
  \belowdisplayskip \abovedisplayskip
  \abovedisplayshortskip 2pt plus1pt  minus1pt%
  \belowdisplayshortskip 2pt plus1pt minus1pt%
}
\g@addto@macro\small{%
  \abovedisplayskip 2pt plus 1pt minus 1pt%
  \belowdisplayskip \abovedisplayskip
  \abovedisplayshortskip 2pt plus1pt  minus1pt%
  \belowdisplayshortskip 2pt plus1pt minus1pt%
}
\g@addto@macro\footnotesize{%
  \abovedisplayskip 2pt plus 1pt minus 1pt%
  \belowdisplayskip \abovedisplayskip
  \abovedisplayshortskip 2pt plus1pt  minus1pt%
  \belowdisplayshortskip 2pt plus1pt minus1pt%
}
\makeatother

\title{VDialogUE: A Unified Evaluation Benchmark \\ for Visually-grounded Dialogue}

\author{Yunshui Li$^{1,2}$ \quad  Binyuan Hui$^{3}$ \quad Zhichao Yin$^{1,4}$ \quad Wanwei He $^{1}$ \quad  Run Luo$^{1}$ \\ \textbf{Yuxing long}$^{3}$  \quad \textbf{Min Yang}$^{1}$\footnotemark[1] \quad \textbf{Fei Huang}$^{3}$ \ \  \textbf{Yongbin Li}$^{3}$\footnotemark[1] \\
        $^{1}$Shenzhen Institute of Advanced Technology, Chinese Academy of Sciences \\
        $^{2}$University of Chinese Academy of Sciences \\
        $^{3}$DAMO Academy, Alibaba Group\\
        $^{4}$University of Science and Technology of China\\
        \text{\{ys.li, min.yang\}@siat.ac.cn, \{binyuan.hby, shuide.lyb\}@alibaba-inc.com}\\
        }

\date{}

\begin{document}
\maketitle
\renewcommand{\thefootnote}{\fnsymbol{footnote}}
\footnotetext[1]{Corresponding author}
\renewcommand{\thefootnote}{\arabic{footnote}}
\begin{abstract}
Visually-grounded dialog systems, which integrate multiple modes of communication such as text and visual inputs, have become an increasingly popular area of investigation.
However, the absence of a standardized evaluation framework poses a challenge in assessing the development of this field.
To this end, we propose \textbf{VDialogUE}, a \textbf{V}isually-grounded \textbf{Dialog}ue benchmark for \textbf{U}nified \textbf{E}valuation. It defines five core multi-modal dialogue tasks and covers six datasets. 
Furthermore, in order to provide a comprehensive assessment of the model's performance across all tasks, we developed a novel evaluation metric called VDscore, which is based on the Analytic Hierarchy Process~(AHP) method.
Additionally, we present a straightforward yet efficient baseline model, named \textbf{VISIT}~(\textbf{VIS}ually-grounded d\textbf{I}alog \textbf{T}ransformer), to promote the advancement of general multi-modal dialogue systems. It progressively builds its multi-modal foundation and dialogue capability via a two-stage pre-training strategy.
 We believe that the VDialogUE benchmark, along with the evaluation scripts and our baseline models, will accelerate the development of visually-grounded dialog systems and lead to the development of more sophisticated and effective pre-trained models.\footnote{\url{https://github.com/AlibabaResearch/DAMO-ConvAI/tree/main/vdialog}}
\end{abstract}

\section{Introduction}
In recent years, there has been a growing interest in visually-grounded dialogue systems, which involve a machine interacting with a human user using natural language and referencing multi-modal contexts like images or videos.~\cite{li2023pace, strub2017end}.
With access to the visual contents beyond text, the dialog systems could perceive the world around them and be able to communicate with humans in a more engaging and efficient way, steering in the next generation of general multi-modal intelligence.
\begin{figure}[t]
  \centering
  \includegraphics[width=\linewidth]{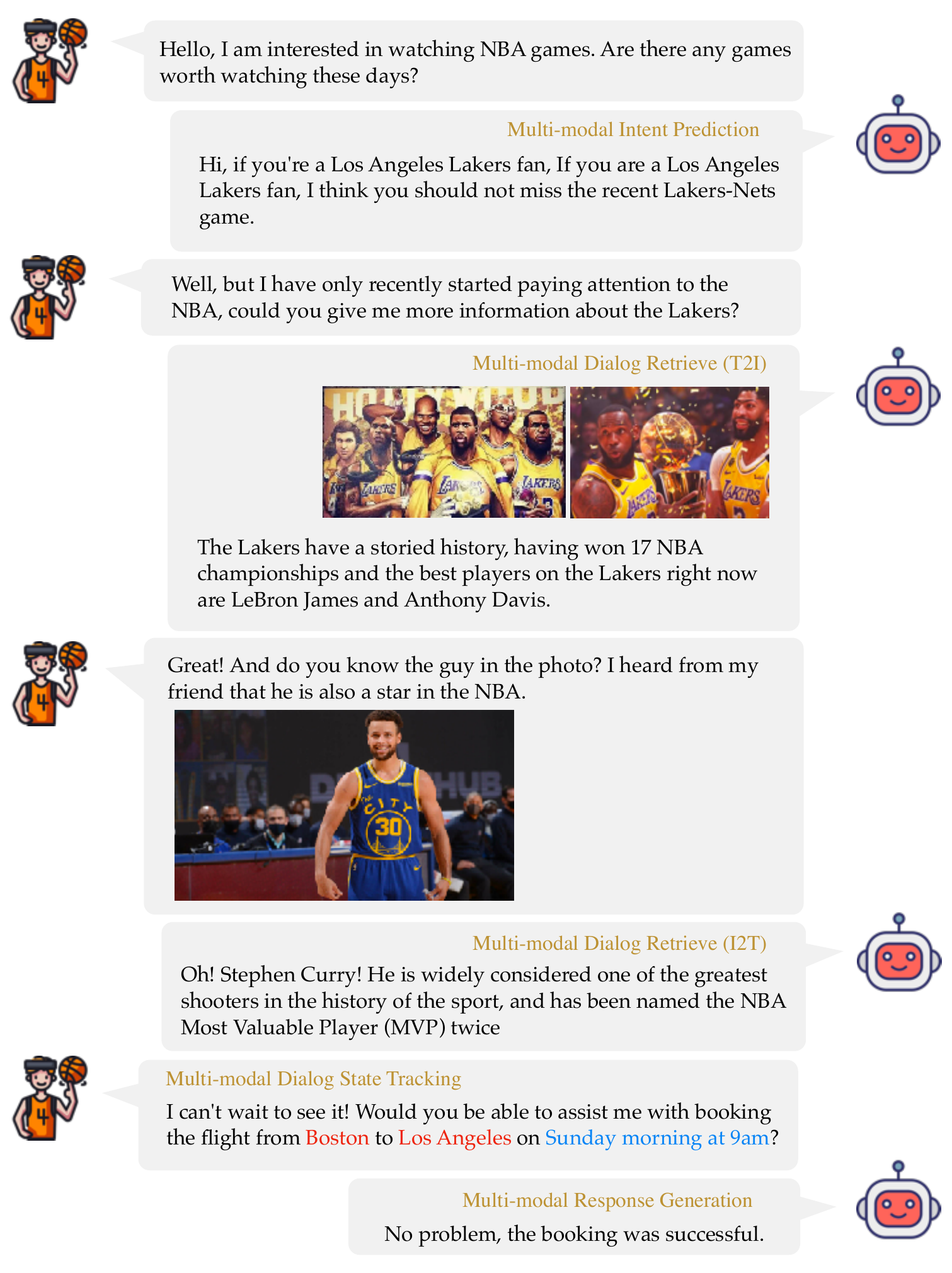}
  \caption{An example of multi-modal dialogue, which involves multiple tasks, including multi-modal intent prediction, multi-modal state tracking, multi-modal
dialog retrieval (T2I\&I2T) and response generation.}
  \label{fig:intro}
\end{figure}

To enhance the development of the visually-grounded dialog systems, a great number of research studies have been conducted in the field of multi-modal dialog datasets, which could be divided into goal-driven dialog (e.g. reserving a restaurant for a user) \cite{young2013pomdp,saha2018towards,liao2021mmconv,kottur2021simmc} and goal-free dialog (e.g., casual ‘chit-chat’ with chatbots) \cite{das2017visual,adiwardana2020towards,mostafazadeh2017image,zang2021photochat,zheng2021mmchat,feng2022mmdialog}.
However, inconsistent evaluation methods across these datasets make it difficult to accurately assess advancements and compare methods with prior work. Even though ChatGPT~\footnote{https://chat.openai.com/} or GPT-4 are often considered a multi-modal dialogue system, there is currently a lack of widely accepted multi-modal dialogue benchmark to evaluate its performance.

Furthermore, different multi-modal dialogue tasks often call for different evaluation metrics, and the importance of various tasks may vary for a general multi-modal dialogue system. Hence, creating a metric for multi-modal dialogue tasks that enables comprehensive and convenient evaluation has become a challenge in the development of multi-modal dialogue systems.

Additionally, it is worth noting that most current visually-grounded dialog models are tailored for one specific task and struggle with out-of-domain data.
Nevertheless, one of the ultimate goals of the multi-modal conversational assistant is capable of performing various dialog tasks grounded in the multi-modal context.


To tackle the above challenge and foster research in this direction, we present \textbf{VDialogUE}, 
 a \textbf{V}isually-grounded \textbf{Dialog}ue benchmark for \textbf{U}nified \textbf{E}valuation, aiming to promote unified multi-modal conversation models that could perform various dialog tasks in different scenarios towards general and modern intelligent assistants. 
Concretely, VDialogUE is the first public multi-task benchmark for visually-grounded dialog systems, which covers six distinct datasets belonging to five fundamental dialog tasks including multi-modal intent prediction, multi-modal dialog retrieval (text-to-image), multi-modal dialog retrieval (image-to-text), multi-modal dialog state tracking and multi-modal response generation. We have also developed a comprehensive evaluation metric named \textbf{VDscore}. The design of VDscore is based on the widely used AHP~\cite{vaidya2006analytic} method, which is a hierarchical analysis method commonly utilized as a quantitative analysis tool to establish the relative importance of various factors in complex problems.
In addition, we provide standardized evaluation scripts and a timely updated leaderboard for fair and easy comparison of visually-grounded dialog systems across different tasks in VDialogUE.

Along with VDialogUE, we release a competitive baseline model called \textbf{VISIT} as the starting point for a general-purpose multi-modal dialogue model.
VISIT is a pre-trained multi-modal conversation model, which can be effectively applied to various downstream visually-grounded dialog tasks. 
To alleviate the issue of limited pre-training data for multi-modal dialogue, we adopt a two-phase training method to pre-train VISIT. In the first phase, we extensively train VISIT on non-dialogue text-image pairs to enhance its multi-modal capabilities at a large scale. In the second phase, we utilize our visually-grounded dialogue corpus, VDialogUE, to further improve its dialogue capabilities.
Experimental results show that VISIT substantially outperforms comparable models trained on each task of VDialogUE separately.
However, our baseline model still achieves a fairly low absolute score on VDscore, which verities the necessity of VDialogUE and developing more sophisticated visually-grounded dialog systems with our unified evaluation benchmark.

In summary, our main contributions are four-fold:
\begin{itemize}
    \item To the best of our knowledge, VDialogUE is the first unified evaluation benchmark for visually-grounded dialogs, comprising six datasets from five core multi-modal dialog tasks.
    \item We have designed the VDscore metric for the comprehensive and convenient evaluation of general-purpose multimodal dialogue models.
    \item Extensive experimentation indicates that VISIT achieves competitive performance compared to strong baselines across various multi-modal dialog tasks.
    \item Our investigation of a two-stage pre-training strategy has demonstrated that it is an effective and efficient method for incremental learning of large models.
\end{itemize}

\section{Related Work}
\subsection{Benchmark Development}
To a certain extent, the emergence of unified benchmarks has driven the development of general models and provided a fair platform for subsequent work comparisons.
GLUE~\cite{wang2018glue} and SuperGLUE~\cite{wang2019superglue} have suggested a unified evaluation framework that simplifies the evaluation of pre-trained language models across various tasks. Additionally, DialoGLUE~\cite{mehri2020dialoglue} and dodecaDialogue~\cite{shuster2019dialogue} offer a benchmark designed specifically for assessing dialogue systems.

Moreover, VALUE~\cite{cao2020behind} proposes probing tasks to delve into the inner workings of vision-language pre-training models, while~\citet{li2021value} develops a video-and-language-focused VALUE benchmark. GEM~\cite{su2021gem} benchmark evaluates multilingual multi-modal models, including image-text retrieval and image captioning. Recently,~\citet{zhou2022vlue} proposed the VLUE benchmark, which is a multi-task and multi-dimensional assessment of vision-language pre-training~(VLP) models.

Although progress has been made in multi-modal and dialogue systems, the exploration of a unified multi-modal dialogue system remains relatively unexplored, despite its significant research value. To fill this gap, we suggest a unified benchmark for visually-grounded dialogue evaluation.

\subsection{Multi-Modal Dialogue Datasets}
In recent years, there has been a proliferation of multi-modal dialogue datasets. VisDial~\cite{das2017visual} is one such dataset where workers generate conversation about a shared image grounded in corresponding images. IGC~\cite{mostafazadeh2017image} is more realistic, but its limited size makes it challenging for model training. Image-Chat~\cite{shuster2018image} is a larger dataset consisting of 202K image-grounded dialogues. PhotoChat~\cite{zang2021photochat} is the first human-human multi-modal dialogue dataset from a real conversation scenario, 
while MMChat~\cite{zheng2021mmchat} features a large-scale Chinese multi-modal dialogue. MMDialog~\cite{feng2022mmdialog} is a million-scale multi-turn dialogue dataset.
Multi-modal datasets also include task-oriented conversation datasets, such as MMD~\cite{saha2018towards} with over 150K fashion domain conversation sessions between shoppers and sales agents, SIMMC$_{1.0}$\cite{moon2020situated} with 13K dialogues on furniture and fashion domains, SIMMC$_{2.0}$\cite{kottur2021simmc} with more complex interaction scenarios than SIMMC$_{1.0}$, and MMConv~\cite{liao2021mmconv} with fully annotated role-playing dialogues covering multiple domains and tasks related to traveling scenarios.

\subsection{Multi-Modal Dialogue Models}
Based on the aforementioned multi-modal dialogue datasets, numerous advanced works have been proposed and developed. 
Some modeling works, such as those by ~\citet{niu2019recursive}, ~\citet{gan2019multi} and~\citet{qi2020two} have been conducted to improve the performance of conversational agents in image-grounded dialogue. Besides, ~\citet{zang2021photochat} proposed a dual-encoder model that utilizes object labels to encode image features in order to perform a dialogue-based image retrieval task. 
Later on, \citet{yang2021open} and \citet{chen2021learning} enhanced the textual expressions of generated dialogue responses through associative vision scenes. \citet{zheng2021mmchat} proposed a multi-modal dialogue generation model based on Seq2Seq architecture. ~\citet{lee2022learning} created a multi-modal encoder-decoder that incorporates visual inputs and performs all sub-tasks via joint learning. More recently, ~\citet{sun2021multimodal} proposed the first multi-modal dialogue response generation model that understands multi-modal contexts and produces informative text and image responses.
As previously mentioned, existing models have achieved success in specific sub-tasks within a given dataset, but they may not be sufficient for addressing diverse multi-modal dialogue tasks. 

\section{VD\MakeLowercase{ialog}UE}
VDialogUE is a multi-task multi-domain visually-grounded dialog benchmark with the goal of providing an accessible platform and a standard practice for the evaluation of general visually-grounded dialog systems. 
As shown in Figure~\ref{fig:intro} and Table~\ref{tab:dialog}, VDialogUE consists of six different datasets spanning over five different tasks: Multi-Modal Intent Prediction, Multi-Modal Dialog Retrieval~(T2I), Multi-Modal Dialog Retrieval~(I2T), Multi-Modal Dialog State Tracking and Multi-Modal Response Generation, where T2I and I2T are short for text-to-image and image-to-text, respectively. Next, we elaborate on the five dialog tasks and six datasets in our VDialogUE benchmark. Ultimately, we present the methodology behind the construction of the VDscore.
\begin{center}
\begin{table*}[t]
\resizebox{\textwidth}{!}{
  \begin{tabular}{lclllllllllc}
    \multicolumn{2}{l}{\SmallImage{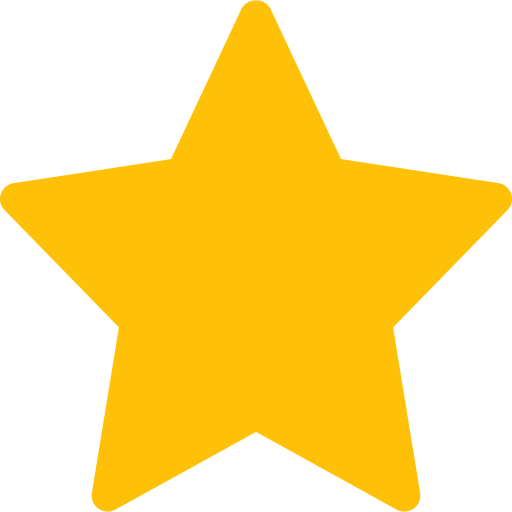} Multi-Modal Intent Prediction}
    & \multicolumn{4}{l}{\SmallImage{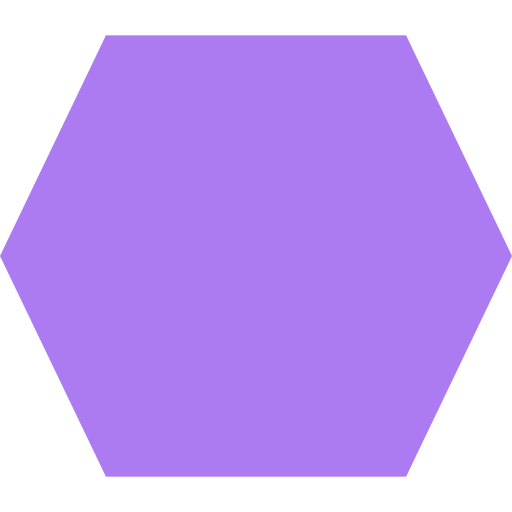} Multi-Modal Dialog Retrieval~(T2I)}
    & \multicolumn{4}{c}{\SmallImage{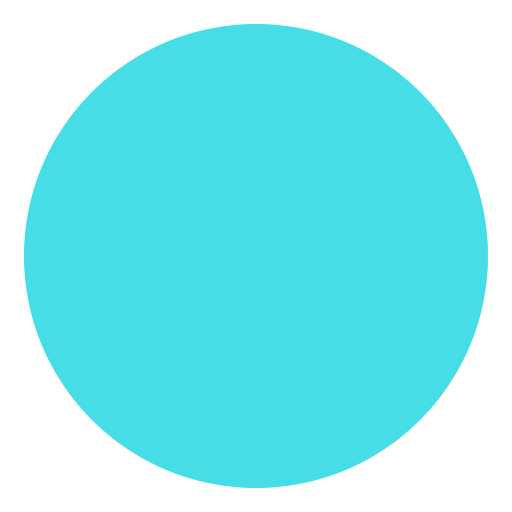} Multi-Modal Dialog Retrieval~(I2T)}& \multicolumn{2}{c}{ }\\
    \multicolumn{2}{l}{\SmallImage{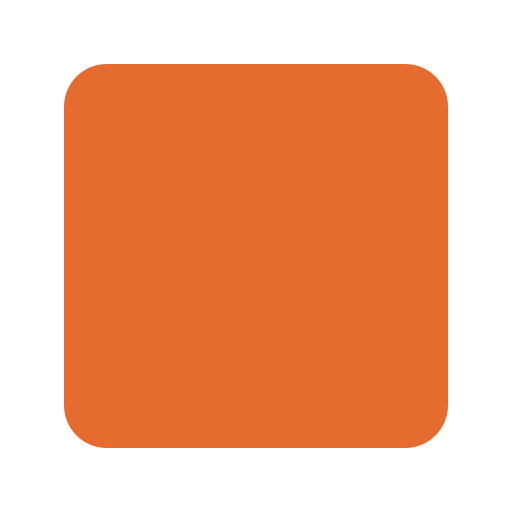} Multi-Modal Dialog State Tracking}
    &\multicolumn{4}{l}{\SmallImage{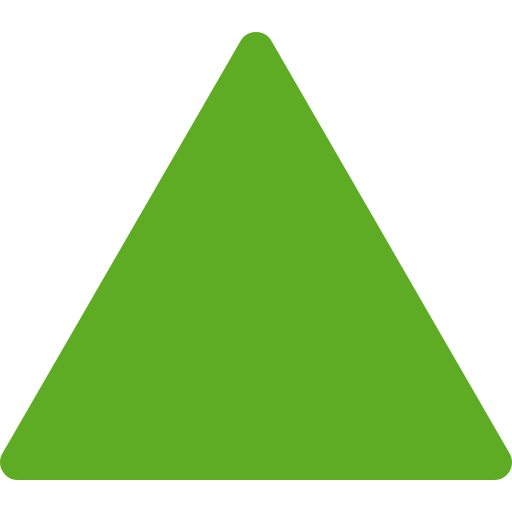} Multi-Modal Response Generation} & \multicolumn{6}{c}{ }\\
    \toprule
    \multirow{2}{*}{Dataset} & \multirow{2}{*}{Tasks} & \multicolumn{3}{c}{Dialogues} & \multicolumn{3}{c}{Turns} & \multicolumn{3}{c}{Images} & \multirow{2}{*}{Task-Oriented}\\

    \cmidrule(lr){3-5} \cmidrule(lr){6-8} \cmidrule(lr){9-11}
    
     & & Train & Dev & Test & Train & Dev & Test & Train & Dev & Test & \\
    
    \midrule
    ImageChat~\cite{shuster2018image} & \SmallImage{fig/circle} & 186,782 & 5,000 & 9,997 & 355,862 & 15,000 & 29,991 & 186,782 & 5,000 & 9,997 &\BigImage{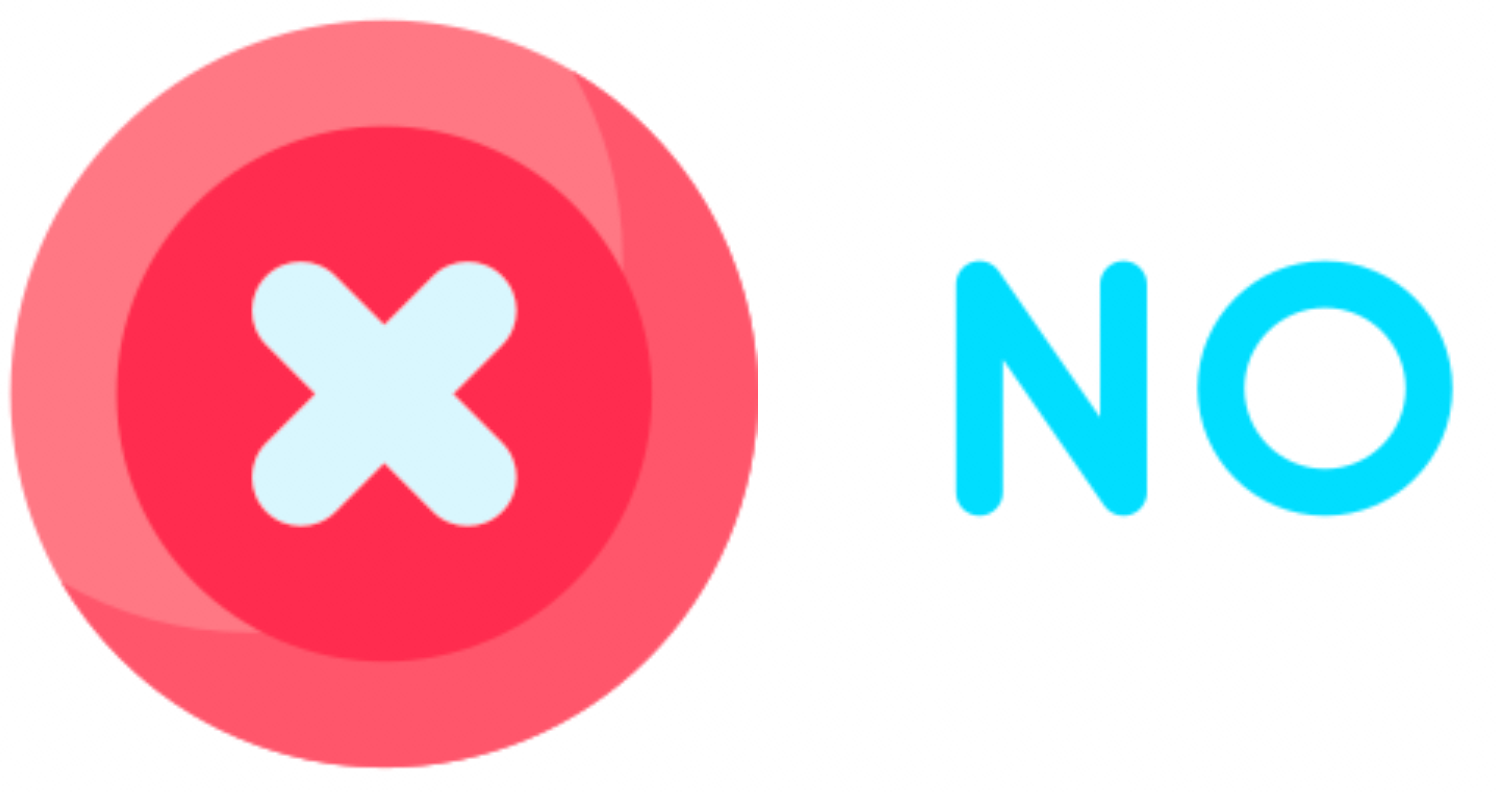} \\
    VisDial$_{1.0}$~\cite{das2017visual} & \SmallImage{fig/circle} & 123,287 & 2,064 & 8,000 & 1232,870 & 20,640 & 8,000 & 123,287 & 2,064 & 8,000 &\BigImage{fig/no} \\
    
    PhotoChat~\cite{zang2021photochat} & \SmallImage{fig/star} \& \SmallImage{fig/hexagon} & 10,286 & 1,000 & 1,000 & 97,586 & 9,533 & 9,590 & 8,917 & 1,000 & 1,000 & \BigImage{fig/no} \\
    MMDialog~\cite{feng2022mmdialog} & \SmallImage{fig/star} \& \SmallImage{fig/hexagon} &1,059,117 & 10,000&  10,000& 4,825,054 &45,382 & 45,798 & 1,509,288& 23,812&  23,766 &\BigImage{fig/no} \\
    
    MMConv~\cite{liao2021mmconv} & \SmallImage{fig/square} \& \SmallImage{fig/triangle} &
    3,500 & 606 & 1,000 & 26,869 & 4,931 & 7,959 & 23,303 & 5,008 & 7,617 & \BigImage{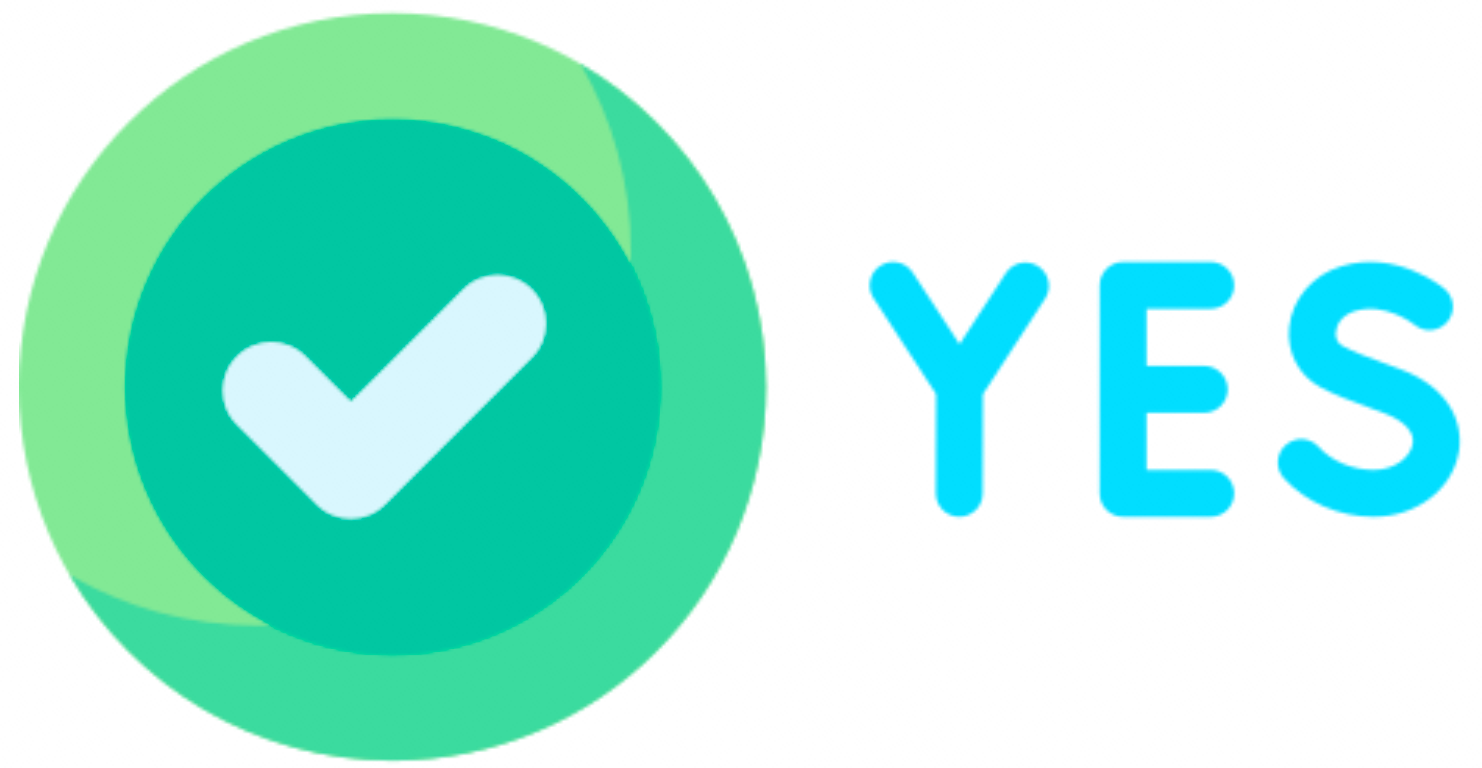}\\
    SIMMC$_{2.0}$~\cite{kottur2021simmc} & \SmallImage{fig/square} \& \SmallImage{fig/triangle} &
    7,307 & 563 & 1,687 & 38,127 & 3,494 & 8,609 & 8,563 & 762 & 1,889 & \BigImage{fig/yes}\\
    \midrule
  \bottomrule
 
\end{tabular}}
  \caption{Characteristics of six datasets in VDialogUE}
  \label{tab:dialog}
\end{table*}
\end{center}

\subsection{Notations}
Given a set of $n$ multi-modal dialogue samples $\mathcal{D}=\left\{\left(U_i, R_i\right)\right\}_{i=1}^n$, where $U_i$ and $R_i$ represent the dialogue context and response, respectively.
Compared to traditional textual dialogue, both $U_i=\{u_k^m\}_{k=1}^K$ and $R_i=\{r_q^m\}_{q=1}^Q$ can incorporate various types of information including textual texts and visual images, where $K$ and $Q$ are the number of elements, and $m \in \{\mathfrak{t}, \mathfrak{v}\}$ denotes the modality of  $U_i$ (or $R_i$). The notation $\mathfrak{t}$ indicates textual utterances, while $\mathfrak{v}$ indicates visual images. 
\subsection{Multi-Modal Intent Prediction}
The aim of the Multi-Modal Intent Prediction task is to identify the specific intent of the user in the multi-modal context. More specifically, it predicts the probability of photo sharing in the upcoming conversation turn. In equation, it’s formulated as a binary classification task:
\begin{equation}
\forall j \in[1, K],\  \mathcal{M}\left(u_{\leq j}, r_{<j}\right) \in\{0,1\},
\end{equation}
where $\mathcal{M}(\cdot, \cdot)$ is the intent prediction model taking the dialogue context $u_{\leq j}$ and response elements $r_{<j}$ of all the previous turns as the input and outputs a binary value. It should predict 1 when photo sharing behavior occurs in the next conversation turn and otherwise 0. 
Note that whether the model make use of all the previous turns is contingent upon the design of the model itself. We use \textit{F1 score}, \textit{precision}, and \textit{recall} as the evaluation metrics for this task following~\cite{zang2021photochat}. 
We describe two different datasets for the task of intent prediction as below.

\paragraph{\textbf{PhotoChat}} is composed of 10,917 distinct images and 12,286 dialogues between humans. The average number of turns per dialogue in PhotoChat is 9.5, and each dialogue is only associated with one image. Moreover, the image is shared, on average, after 7 turns of conversation. However, a major issue with PhotoChat is its imbalanced class distribution for multi-modal intent prediction, as there are more negative examples than positive ones. In addition, the dataset contains images that are reused across multiple dialogues, as shown in Table~\ref{tab:dialog}. 

\paragraph{\textbf{MMDialog}} is a comprehensive open-domain multi-modal dialogue dataset, containing a carefully selected set of 1.08 million authentic dialogues and 1.53 million distinct images covering 4,184 topics. Typically, each dialogue session in MMDialog contains 2.59 images and 4.56 turns, with the images placed at any point in the conversation. This ensures a relatively even distribution of positive and negative samples. Additionally, each turn in MMDialog contains an average of 16.64 text tokens, which is significantly higher than the average of 6.33 text tokens per turn in PhotoChat.

\subsection{Multi-Modal Dialog Retrieval~(T2I)}
This task requires the model to retrieve the most relevant image from a candidate image set $\left\{r_z^{\mathfrak{v}}\right\}_{z=1}^Z$ while given the dialog history, where $Z$ is the size of the set.
It requires models to extract key object information from the conversation and exclude irrelevant distractors. Additionally, the model needs to be capable of aligning the semantic space of visual and linguistic modalities, such that two perspectives of a scene are similarly represented in the vector space.
Following~\cite{zang2021photochat}, we use $Recall@K(R@K)$, computed as “the fraction of times a correct item was found among the top K results” as the evaluation metrics.

We select \textbf{PhotoChat} and \textbf{MMDialog} datasets for multi-modal dialog retrieval~(T2I) task in the VDialogUE benchmark. These two datasets have some differences in the selection of candidate image sets for evaluation. Since PhotoChat has a small number of test set images, the entire test set images are used as the candidate set for each dialogue. On the other hand, MMDialog has a large number of test set images (23k), making it difficult to use the entire test set as the image candidate set. Therefore, the developers manually selected 999 images as negative samples for each dialogue in the test set. As a result, the candidate set size for both datasets is 1,000. 
\subsection{Multi-Modal Dialog Retrieval~(I2T)} 
It requires the model to take as input an image and a multi-turn, free-form, open-ended, natural language question about the image and produces or selects a natural language answer as the output. As with previous work, the task is defined to answer the question by selecting the most compatible target sample from the text candidate answers $\left\{r_z^t\right\}_{z=1}^Z$ in VDialogUE, where $Z$ is the size of the set. 
The candidate answers contain both ground truth samples and indistinguishable negative samples, and the model is asked to return a sorting of the candidate answers. The performance for Multi-Modal Dialog Retrieval~(I2T) is measured by $R@K$. We introduce two different datasets for this task.

\paragraph{\textbf{VisDial$_{1.0}$}} contains 123k dialogues. Each dialogue consists of an image and ten rounds of question and answer pairs. The entire discussion centers on the image. Specially, each dialog contains a list of $Z = 100 $ candidate answers at test time. However, answers in VisDial only have 2.9 words in mean length. 

\paragraph{\textbf{ImageChat}} is a dataset of grounded human-human conversations, where speakers are asked to play roles given a provided emotional mood or style. It consists of 202k diverse images and 401k utterances over the images, with 215 different style traits (e.g., optimistic, skeptical or frivolous) to promote engaging conversation. Unlike VisDial$_{1.0}$, the utterances in ImageChat are closer to a normal conversation length, with an average of 12.4. The candidate set size is 100 during evaluation as the same as VisDial$_{1.0}$.

\subsection{Multi-Modal Dialog State Tracking} 
Multi-modal dialog state tracking~(MMDST) extend the traditional notion of the textual dialog state tracking, where pre-defined slots $S=\left\{S_1, \ldots, S_N\right\}$ are grounded on the multi-modal context. The model $\mathcal{M}$  determines for every turn whether any of the slot pairs in present and to predict the slot values. We select MMConv and SIMMC$_{2.0}$ datasets for MMDST in VDialogUE benchmark. $Accuracy$ is used for MMConv evaluation and the SIMMC$_{2.0}$'s performance is measured by the joint \textit{F1} for the cumulative intent, slot predictions.

\paragraph{\textbf{MMConv}} is a task-oriented dataset collected by enabling multi-modal conversations between human-to-human role-playing pairs under real life travel scenarios. The MMConv dataset consists of 751 single-modality dialogues and 4,355 multi-modality dialogues, respectively. MMConv splits MMDST into two tracking subtasks, i.e. categorical and non-categorical tracking. For the categorical one, it selects the most plausible values from the pick-lists based on the contextual representation. For the non-categorical one, it needs to find text spans to fill in the corresponding slots. 

\paragraph{\textbf{SIMMC$_{2.0}$}} includes a total of 11.2k task-oriented dialogs between user and assistant in shopping domain, split into 7.2k and 4k dialogs from fashion and furniture domains respectively. It noted that SIMMC$_{2.0}$ use photo-realistic, virtual renderings of cluttered shopping environments to replicate real-world settings. It requires the model to predict slot and intent information for each dialog. 
\subsection{Multi-Modal Response Generation} 
Given the multi-modal dialogue context $U$, the multi-modal response generation task aims to generate the textual response $R$ by modeling $\boldsymbol{P}(R \mid U ;\theta)$, where $\theta$ is the model $\mathcal{M}$ parameters.
We select \textbf{MMConv} and \textbf{SIMMC$_{2.0}$} datasets for multi-modal response generation task in the VDialogUE benchmark. 
We use the widely used \textit{BLEU}~\cite{papineni2002bleu} to measure the response generation quality.
\begin{table}[t]
\centering
  \resizebox{0.45\textwidth}{!}{
  \begin{tabular}{lccccc}
    \toprule
    \textbf{Task}  &\SmallImage{fig/triangle}  & \SmallImage{fig/star}& \SmallImage{fig/hexagon} & \SmallImage{fig/circle} &\SmallImage{fig/square} \\
    \midrule
    \SmallImage{fig/triangle}~MM Response Generation   & 1 &  2 & 3 & 3 & 5  \\
    \SmallImage{fig/star}~MM Intent Prediction    & 1/2  &  1 & 2& 2 & 4  \\
    \SmallImage{fig/hexagon}~MM Dialog Retrieval (T2I)   & 1/3 & 1/2  & 1& 1 & 3\\
    \SmallImage{fig/circle}~MM Dialog Retrieval (I2T)  & 1/3 & 1/2 &  1& 1 & 3 \\
    \SmallImage{fig/square}~MM Dialog State Track  & 1/5 & 1/4 & 1/3 &1/3 & 1\\
  \bottomrule
\end{tabular}}
\caption{Pairwise comparison matrix of five core multi-modal dialog tasks. For example, \SmallImage{fig/square} vs \SmallImage{fig/triangle}(1/5) shows that \SmallImage{fig/triangle} is more important than \SmallImage{fig/square}. }
\label{tab:vdscore}
\end{table}
\subsection{VDscore}
With the aim of constructing a unified multi-modal dialogue system, we utilized AHP method (Analytic Hierarchy Process), a quantitative analysis tool, to determine the relative importance of different tasks. The most crucial demand for multi-modal dialogue systems is to produce responses that are coherent and contextually relevant, while intent recognition is considered a prerequisite for retrieval tasks, and dialogue tracking is just an intermediate task in continuous dialogue. Based on this assumptions, we created a pairwise comparison matrix as shown in Table~\ref{tab:vdscore}. Then, we calculated the corresponding weights for each task and performed a consistency check to validate our assumptions. More details are stated in Appendix~\ref{detail_vdscore}. The final score of VDscore is calculated as: 
\begin{small}
\begin{equation}
\begin{split}
    \boldsymbol{VDscore} &= \boldsymbol{0.41}\cdot \TinyImage{fig/triangle}
     + \boldsymbol{0.25}\cdot \TinyImage{fig/star} \\&+ \boldsymbol{0.14}\cdot \TinyImage{fig/circle}
     + \boldsymbol{0.14}\cdot \TinyImage{fig/hexagon} + \boldsymbol{0.06} \cdot \TinyImage{fig/square} \\ 
\end{split}
\end{equation}
\end{small}
where each icon represents the average measure of a specific task, obtained by averaging internal metrics and then averaging across different datasets. 
\begin{figure*}[t]
  \centering
  \includegraphics[width=0.75\textwidth]{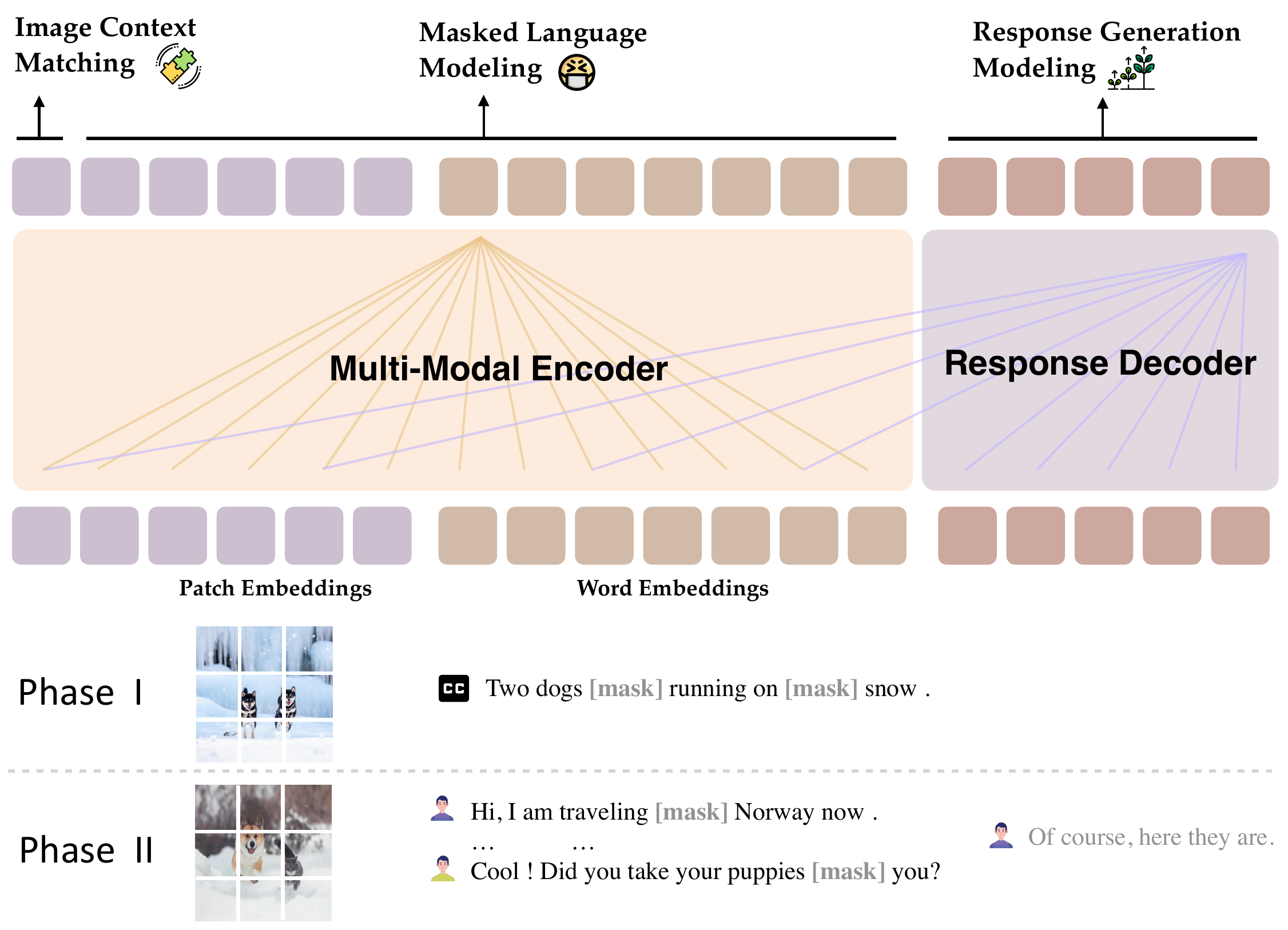}
  \caption{The VISIT model architecture consists of a multi-modal encoder and a response decoder, with different self-attention positions denoted by lines of different colors in each module. The encoder is bi-directional, while the decoder is uni-directional. During Phase-I, the model is trained on multi-modal non-dialog data without response generation modeling. In Phase-II, three pre-training objectives are carried out to optimize the VISIT model jointly on multi-modal dialog data.}
  \label{fig:model}
\end{figure*}

\section{VISIT}

\subsection{Architecture}
As illustrated in Figure~\ref{fig:model},
VISIT has a succinct architecture as a baseline model for VDialogUE, which adopt the standard Transformer as modality interaction backbone and the simplest visual and text embedding scheme.
\paragraph{\textbf{Visual Embedder}}
To minimize overhead, we adopt the patch projection embedding introduced by ViT~\cite{dosovitskiy2020image}. Formally, we process the visual image ${v}\in \mathbb{R}^{H \times W \times {C}}$ by dividing it into $N=H W / P^2$ patches and flattened to ${v}^p \in \mathbb{R}^{N \times\left(P^2 {C}\right)}$, where ${C}$ is the number of channels, $(H, W)$ is the resolution of the input image, and $P$ is the patch resolution. The image patches are processed by a linear projection using a weight matrix $\mathbf{W}_{V} \in \mathbb{R}^{(P^2 \cdot C) \times E}$ and a position embedding $\mathbf{W}_{V}^{\text{pos}} \in \mathbb{R}^{(N+1) \times E}$, resulting in patch embedding $\bar{v} \in \mathbb{R}^{N \times E}$, where $E$ is the dimension of embedding. 
The position embedding is used to add additional information about the position of the patch in the image. 
\begin{equation}
\bar{v} = \left[v_{cls};v_1^p \mathbf{W}_{V};\cdot \cdot \cdot ;v_N^p\mathbf{W}_{V} \right] + \mathbf{W_{V}^{\text{pos}}}
\end{equation}
where $v_{cls}$ is the extra learnable embedding of the image.
\paragraph{\textbf{Textual Embedder}}
The input text $t \in \mathbb{R}^{\mathfrak{L} \times |O|}$ is embedded into a dense representation $\bar{t} \in \mathbb{R}^{\mathfrak{L} \times E}$ by using a word embedding matrix $\mathbf{W}_T \in \mathbb{R}^{|O| \times E}$ and a position embedding matrix $\mathbf{W}_T^{\text{pos}} \in \mathbb{R}^{(\mathfrak{L}+1) \times E}$, where $|O|$ is the size of the vocabulary, $\mathfrak{L}$ is the length of text, and $E$ is the dimension of embedding. It is noteworthy that we usually concatenate the context with the current utterance to form the final textual input. 
\begin{equation}
\bar{t} = \left[t_{cls};t_1 \mathbf{W}_{T};\cdot \cdot \cdot ;t_{\mathfrak{L}} \mathbf{W}_{T} \right] + \mathbf{W_{T}^{\text{pos}}}
\end{equation}
where $t_{cls}$ is the extra learnable embedding of the text.
\paragraph{\textbf{Backbone Network}}
The backbone of \ VISIT consists of $L$ stacked blocks that include a multiheaded self-attention(\textbf{MSA}) layer and an \textbf{MLP} layer. Similar to ViT~\cite{dosovitskiy2020image}, the position of layer normalization~(\textbf{LN}) comes before MSA in VISIT. 

The image and text embeddings are summed with their corresponding modal-type embedding vectors $v^{type}, t^{type} \in \mathbb{R}^{E}$ to achieve the following input to the backbone network:
\begin{equation}
\boldsymbol{H}_0 = \left[ \bar{v}+v^{type} ; \bar{t}+t^{type} \right] 
\end{equation}
The contextualized vector $\boldsymbol{H}_0$ is iteratively updated through L-depth transformer layers up until the final contextualized sequence $\boldsymbol{H}_L$. 
\begin{equation}
\begin{array}{cc}
     \boldsymbol{H}'_l = \boldsymbol{MSA}(\boldsymbol{LN}(\boldsymbol{H}_{l-1})) + \boldsymbol{H}_{l-1}  \\
     \boldsymbol{H}_l = \boldsymbol{MLP} (\boldsymbol{LN}(\boldsymbol{H}'_l)) +\boldsymbol{H}'_l 
\end{array}
\end{equation}
where $l \in \left[ 1, L\right]$. 
Then the first index of sequence $\boldsymbol{H}_L$ followed by linear projection $\mathbf{W}_{pool}\in \mathbb{R}^{E \times E} $ and hyperbolic tangent function, we obtain a pooled representation of the whole multi-modal input $\mathfrak{p}$.
\begin{table}[t]
\centering
  \resizebox{0.5\textwidth}{!}{
  \begin{tabular}{lccc}
    \toprule
    \textbf{Dataset}  & \textbf{Images} & \textbf{Captions} & \textbf{Len}\\
    \midrule
    MSCOCO~\cite{lin2014microsoft}  & 113K &  567K & 11.81  \\
    VG~\cite{krishna2017visual}     & 108K &  5.41M & 5.53  \\
    GCC~\cite{sharma2018conceptual}   & 3.01M& 3.01M  & 10.66\\
    SBU~\cite{ordonez2011im2text}    & 867K & 867K & 15.0 \\
  \bottomrule
\end{tabular}}
\caption{Dataset statistics of multi-modal non-dialogue data. Len is the average token length from  bert-base-uncased tokenizer}
\label{tab:nondial}
\end{table}

\subsection{Two-Phase Pre-training}
As shown in Table~\ref{tab:nondial}, the availability of sufficient paired image-text data enables the model to learn the fundamental inter-modal alignment. Additionally, pre-training the model on multi-modal dialogue data enhances its ability to process not just simple text but also complex, context-dependent dialogues.
Therefore, as illustrated in Figure~\ref{fig:model}, we divided the pre-training process into two phases. In phase-I, VISIT is trained on image-text paired data using two standard pre-training objectives~\cite{kim2021vilt}, namely image-text matching~(\textbf{ITM}) and masked language modeling~(\textbf{MLM}).  we integrate response generation modeling~(\textbf{RGM}) into the pre-training process, which builds upon the foundations laid out in phase-I and is trained on multi-modal dialogue data.

\paragraph{\textbf{Image Text Matching}} 
Given a caption or a multi-turn dialogue, we randomly replace the aligned image with a different image with the probability of 0.5. We employ the representation of the pooled output $\mathfrak{p}$ as the input for a binary classification network \textbf{ITM} head to predict the alignment between current text and image. The loss function of \textbf{ITM} is defined as:
\begin{equation}
\mathcal{L}_{\mathrm{itm}}=\mathbb{E}_{(v, t) \sim D} \ CE\left(\boldsymbol{y}_{\mathrm{itm}}, \boldsymbol{p}_{\mathrm{itm}}(v, t)\right)
\end{equation}
where $D$ can be either multi-modal non-dialogue data $D_n$ or multi-modal dialogue data $D_d$, depending on the specific phase of the training, $CE$ is cross-entropy loss function, $\boldsymbol{y}_{*}$ is ground-truth label and $\boldsymbol{p*}$ is the prediction result of the model, the variables $v$ and $t$ represent the visual image and text, respectively. 

\paragraph{\textbf{Masked Language Modeling}} Using BERT's approach~\cite{devlin2018bert}, we replace random tokens with [MASK] and train the model to predict them based on context and visual cues. We use a 15\% masking probability and feed the output vectors into a two-layer MLP classifier (MLM head) for cross-entropy loss training 

\begin{equation}
\mathcal{L}_{\mathrm{mlm}}=\mathbb{E}_{(v, \hat{t}) \sim D} \ CE\left(\boldsymbol{y}_{\mathrm{mask}}, \boldsymbol{p}_{\mathrm{mask}}(v, \hat{t})\right)
\end{equation}
where $\hat{t}$ is a masked text.

\paragraph{\textbf{Response Generation Modeling}}
The response generation task is performed in an auto-regressive manner, where the appropriate system response $r$ is generated based on the past dialogue history $c$ and associated images $v$. We utilize the standard negative log-likelihood loss $\mathcal{L}_{\mathrm{rgm}}$ for this generation task and implement a similar approach to UniLM~\cite{dong2019unified}.
\begin{equation}
\mathcal{L}_{\mathrm{rgm}}=-\sum_{n=1}^N \log \boldsymbol{p}_{\mathrm{rgm}}\left(r_{n} \mid c, v, r_{<n}\right)
\end{equation}
where $r_n$ is the n-th word in $r$ and $r_{<n} = \{r_1, ..., r_{n−1}\}$ represents the words of previous steps.

In summary, the joint loss for the two phase could be respectively formulated as: 
\begin{equation}
\mathcal{L}^{\mathrm{I}} = \mathcal{L}_{\mathrm{itm}} + \mathcal{L}_{\mathrm{mlm}}
\end{equation}
and for the second phase as:
\begin{equation}
\mathcal{L}^{\mathrm{II}} = \mathcal{L}_{\mathrm{itm}} + \mathcal{L}_{\mathrm{mlm}} + \mathcal{L}_{\mathrm{rgm}}
\end{equation}
\subsection{Fine-Tuning on Downstream Tasks}
\begin{center}
\begin{table*}[t]
    \centering
    \resizebox{0.80\textwidth}{!}{
    
    \begin{tabular}{lllcllllc}
    \toprule
    \textbf{Task} & & \textbf{Dataset} & \ & \textbf{Metric} &\ & \textbf{Previous SOTA} &\   & \textbf{VISIT} \\
    \cmidrule(lr){1-9}
    \cmidrule(lr){1-9}
    \multirow{6}{*}{\SmallImage{fig/star}~Multi-Modal Intent Prediction} &
    & \multirow{3}{*}{PhotoChat} & \ & F1 & \ & 58.9~(T5-3B)& \ &\textbf{60.6}\\
    & & & \ & Precision & \ & 58.2~(T5-base) & \ & \textbf{61.5}\\
    & & & \ & Recall & \ & 64.6~(T5-3B) & \ &  \textbf{66.8}\\
    \cmidrule(lr){3-9}
    
     & \ &  \multirow{3}{*}{MMDialog} & \ & F1 & \ & 75.5 (Divter) & \ & \textbf{76.3}\\
     & & & \ & Precision & \ & 72.3~(ViLT) & \ & \textbf{75.1} \\
     & & & \ & Recall & \ & 76.4~(ViLT) & \ & \textbf{80.9} \\
    \cmidrule(lr){1-9}
    
    \multirow{6}{*}{\SmallImage{fig/hexagon}~Multi-Modal Dialog Retrieval(T2I)}  & & \multirow{3}{*}{PhotoChat} & \ & R@1 & \ & 10.4~(SCAN) & \ & \textbf{13.8}\\
     & & & \ & R@5 & \ & 27.0~(SCAN)& \ &  \textbf{32.7}\\
    & & & \ & R@10 & \ & 37.1~(SCAN) & \ & \textbf{42.3}\\
    \cmidrule(lr){3-9}
     & & \multirow{3}{*}{MMDialog} & \ & R@1 & \ & \textbf{29.6}~(DE++)  & \ & 20.8\\
     & & & \ & R@5 & \ & 45.1~(DE++)  & \ & \textbf{46.0}\\
    &  &  &\ & R@10  & \ & 53.6~(DE++)& \ & \textbf{58.0}\\
    \cmidrule(lr){1-9}
    \multirow{4}{*}{\SmallImage{fig/circle}~Multi-Modal Dialog Retrieval(I2T)}  & & \multirow{2}{*}{Image-Chat} & \ & R@1 & \ & \multicolumn{2}{l}{50.3~(TransResNet)} & \textbf{51.5}\\
     & & & \ & R@5 & \ & \multicolumn{2}{l}{\textbf{75.4}~(TransResNet)} & 73.2\\
    \cmidrule(lr){3-9}
     & & \multirow{2}{*}{VisDial} & \ & R@1 & \ & \multicolumn{2}{l}{\textbf{55.7}~(UTC)}  & 51.4\\
    &  & & \ & R@5 & \ & \multicolumn{2}{l}{\textbf{84.8}~(UTC)} & 82.6\\
    \cmidrule(lr){1-9}
    \multirow{3}{*}{\SmallImage{fig/square}~Multi-Modal Dialog State Tracking} &  & \multirow{2}{*}{SIMMC2.0} & \ & Intent-F1 & \ & \multicolumn{2}{l}{96.3~(BART-large)}  & \textbf{96.7} \\
     & & & \ &  Slot-F1 & \ & \multicolumn{2}{l}{\textbf{88.3}~(BART-large)} & 86.6\\
    \cmidrule(lr){3-9} 
     & &\multirow{1}{*}{MMConv} & \ & Accuracy  & \ & \multicolumn{2}{l}{18.0~(DS-DST)}  & \textbf{32.7} \\
    \cmidrule(lr){1-9}
    \multirow{2}{*}{\SmallImage{fig/triangle}~Multi-Modal Response Generation} &  & SIMMC2.0 & \ & BLEU & \ &\multicolumn{2}{l}{33.1~(BART-large)} & \textbf{33.4}\\
    \cmidrule(lr){3-9}
     & & \multirow{1}{*}{MMConv}  & \ &  BLEU & \ & \multicolumn{2}{l}{20.3~(SimpleTOD)}  & \textbf{21.2} \\%
    \cmidrule(lr){1-9}
    \SmallImage{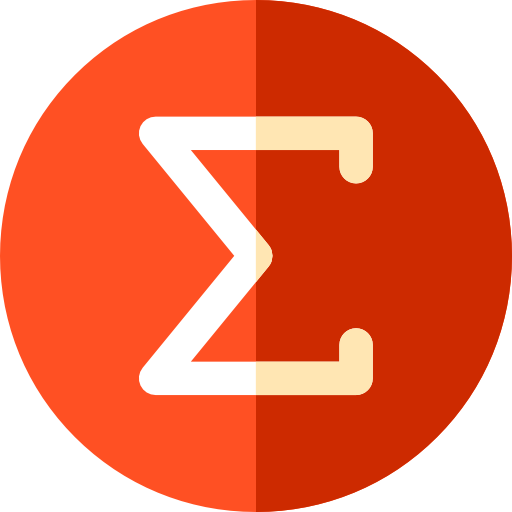}~Overall &  & All of Above & \ & VDscore & \ & \multicolumn{2}{l}{45.2} & \textbf{46.5} \\
    \bottomrule
    \end{tabular}}
    \caption{
    Experimental results on the VDialogUE benchmarks. 
    }
    \label{tab:benchmark}
\end{table*}
\end{center}

Once the pre-training of VISIT is finished, we perform fine-tuning on specific downstream tasks. To tackle the multi-modal intent prediction and dialog retrieval tasks, we start by initializing the similarity score head with the ITM head that was pre-trained, specifically the component responsible for computing the true-pair logits. We then randomly select 15 text samples to act as negative examples and fine-tune the model using cross-entropy loss, with the goal of maximizing scores on positive pairs. To facilitate multi-modal dialog state tracking in the MMConv dataset, we augment the model architecture with a CATEGORICAL head and an SPAN head. The former handles categorical slots, while the latter is responsible for non-categorical ones. For the MMDST task in SIMMC$_{2.0}$ and multi-modal response generation, we fine-tune the model by calculating the standard negative log-likelihood loss $\mathcal{L}_{\mathrm{rgm}}$ in an end-to-end manner.
\section{Experiments}
\subsection{Experimental Setting}
We initialize the Transformer weights with ViT-B/32~\cite{dosovitskiy2020image} pretrained on ImageNet. The embedding size $E$ is 768, layer depth $L$ is 12, patch size $P$ is 32, MLP size is 3,072, and the number of attention heads is 12. We use \textit{bert-base-uncased} tokenizer to tokenize text inputs. Patch projection of VISIT yields $12 \times 20 = 240$ patches for an image with a resolution of $384 \times 384$. For all experiments, we use AdamW optimizer~\cite{loshchilov2017decoupled} with base learning rate $10^{-4}$ and weight decay of $10^{-2}$. The learning rate is warmed up for 10\% of the total training steps and is decayed
linearly to zero for the rest of the training. In the pre-training process, we conduct 200K and 25K steps for each of the two phases respectively with a batch size of 4,096. For all downstream tasks, we train for ten epochs with a batch size of 256.  

\subsection{Baselines}
Since there is currently no general model available to solve all tasks in VDialogUE, we choose the models that exhibit the best performance on specific tasks to serve as our baseline models. We compare VISIT with previous
    state-of-the-art models, including T5~\cite{raffel2020exploring}, Divter~\cite{feng2022mmdialog}, ViLT~\cite{kim2021vilt}, SCAN~\cite{lee2018stacked}, DE++~\cite{feng2022mmdialog}, TransResNet~\cite{shuster2018image}, UTC~\cite{chen2022utc}, BART~\cite{lewis2019bart}, DS-DST~\cite{zhang2019find} and SimpleTOD~\cite{hosseini2020simple}. Appendix~\ref{baselinemodel} contains further descriptions of the baseline models.

\subsection{Overall Performance}
The experimental results on the VDialogUE benchmark are presented in Table~\ref{tab:benchmark}. We observe that our VISIT model can achieve competitive
performance compared to strong baselines for a broad variety of visually-grounded dialogue tasks. More precisely, our VISIT model demonstrated state-of-the-art performance on certain tasks, including multi-modal intent prediction, dialog retrieval(T2I), dialog state tracking and response generation, across almost all metrics. Specifically, we found that our model also achieved consistent improvement in the comprehensive evaluation of VDscore. 
\begin{table}[b]
    \centering
    \resizebox{0.45\textwidth}{!}{
    \begin{tabular}{lllllll}
    \toprule
    \multirow{2}{*}{\textbf{Model}} & \multicolumn{3}{c}{\textbf{PhotoChat}} & \multicolumn{2}{c}{\textbf{Image-Chat}} \\
    \cmidrule(lr){2-4}
    \cmidrule(lr){5-6}
    
     & \textbf{R@1} & \textbf{R@5} & \textbf{R@10}& \textbf{R@1} & \textbf{R@5}\\
    \midrule
    \rowcolor{gray!20} VISIT  & \textbf{13.8} & \textbf{32.7} & \textbf{42.3}&\textbf{51.5} & \textbf{73.2} \\
    VISIT$_{Phase-I}$ & 10.2 & 27.1 & 35.4 & 48.2 & 69.9\\	 
    VISIT$_{Phase-II}$ & 10.5 & 26.4 & 35.0 & 49.1 & 68.4\\
    \bottomrule
    \end{tabular}
    }
 \caption{Ablation test results on the multi-modal dialog retrieval task by using different pre-training data.}
 \label{tab:ab-data}
\end{table}
\subsection{Analysis and Limitation}
However, we noticed that the VISIT model does not perform as well on text retrieval tasks as it does on other tasks, particularly on the VisDial dataset. We speculate that there are several reasons for this. 
Firstly, VisDial shows a clear distribution bias towards image content over real dialogue scenarios, allowing dialogue agents to rely solely on image features and ignore dialogue context~\cite{kim2020modality,le2022multimodal}.
Secondly, there is an annotator bias that can lead to harmful causal links between dialogue context and output response as demonstrated in~\cite{qi2020two}.
Thirdly, in multi-modal text retrieval, the model needs to prioritize candidate answers over the dialogue history, which is often much longer than the candidate answers consisting of only a few characters. However, our model failed to take this into account and treated the dialogue history and answer equally by simply concatenating them as the input on the text side.

\begin{figure}[t]
  \centering
  \includegraphics[width=0.95\linewidth]{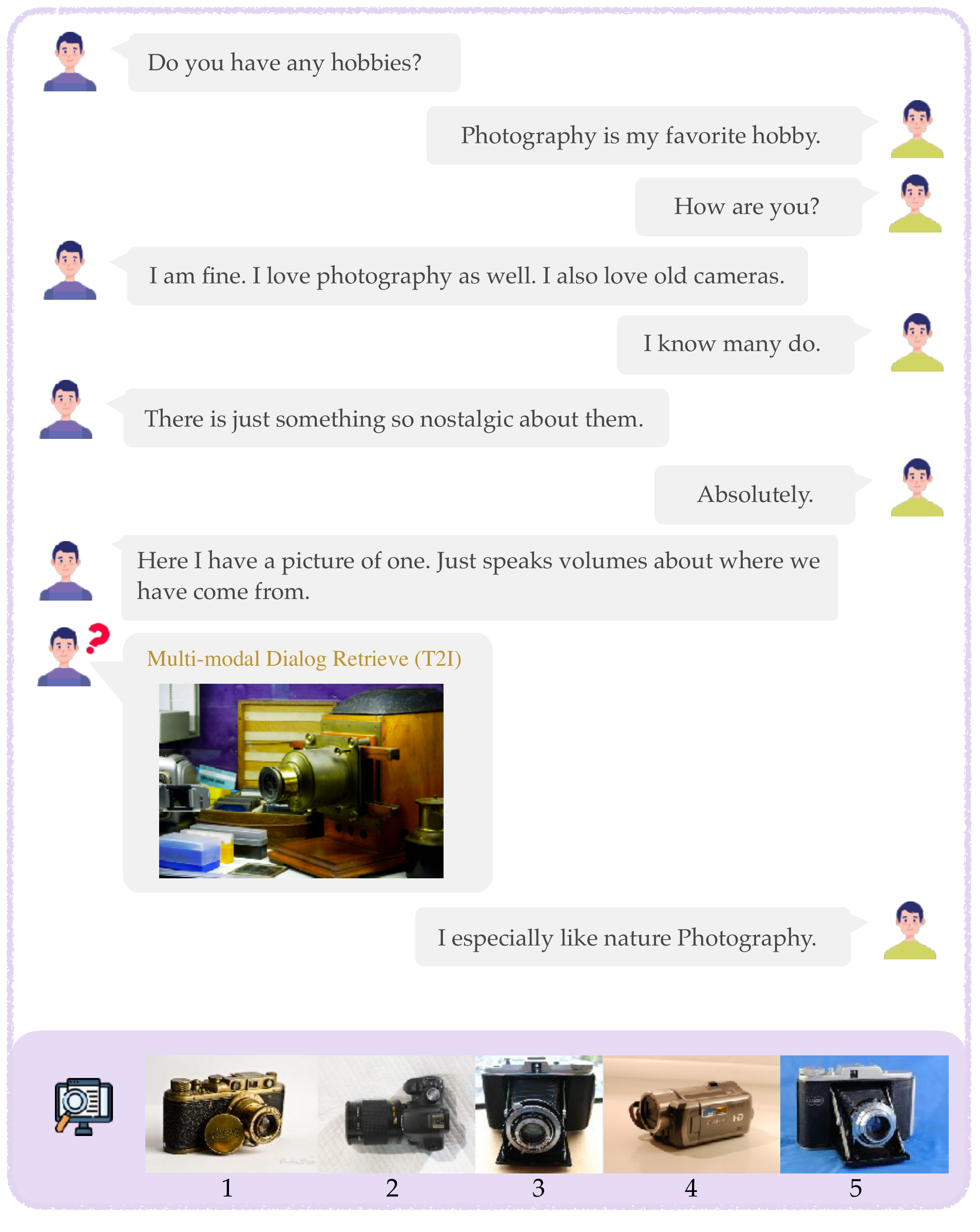}
  \caption{An image retrieval case on the PhotoChat test set. We show the top-5 ranked images from left to right.}
  \label{fig:case}
\end{figure}

In addition, we found that there is still significant room for improvement in the accuracy of our model on image retrieval tasks, particularly on the $R@1$ metric. To better analyze the limitations of VISIT, we carry out an analysis of the errors made by VISIT on the PhotoChat and MMDialog test dataset. As Figure~\ref{fig:case} shows that due to the existence of many similar images in the datasets, VISIT struggles to differentiate some correct images from similar candidates. This limitation may be attributed to the lack of an explicit fine-grained reasoning module that can effectively capture the nuances in both images and texts.

\subsection{Ablation Study}
Furthermore, we perform an ablation study to examine the effect of different pre-training data on both the PhotoChat and Image-Chat datasets. The models that are exclusively pre-trained on multi-modal non-dialog and dialog data are denoted as VISIT$_{Phase-I}$ and VISIT$_{Phase-II}$, respectively. The ablation test results on PhotoChat and Image-Chat are provided in Table~\ref{tab:ab-data}.
It is evident that both the multi-modal non-dialog and dialog pre-training corpora significantly enhance the performance of VISIT. 
This outcome is not surprising as the multi-modal non-dialog data aids the model in acquiring exceptional image-text representations and their alignment, while the multi-modal dialog data stimulates VISIT to capture the contextual information of the dialog.

\section{Conclusion}
The development of visually-grounded dialog systems has gained popularity, but the absence of a standardized evaluation framework presents a challenge for assessing the progress in this field. The proposed VDialogUE benchmark, along with the development of the novel VDscore evaluation metric and the VISIT baseline model, provides a comprehensive assessment of model performance and promotes the advancement of general multi-modal dialogue systems. The VDialogUE benchmark and associated resources are expected to accelerate the development of visually-grounded dialog systems and facilitate the creation of more sophisticated and effective pre-trained models.

\bibliography{tacl2021}
\bibliographystyle{acl_natbib}

\onecolumn

\appendix
\section{Baseline Models}
\label{baselinemodel}
\paragraph{\SmallImage{fig/star}~\textbf{Multi-Modal Intent Prediction}} The T5~\cite{raffel2020exploring} series of models are leading the pack on the PhotoChat dataset. Divter is a strong multi-modal dialogue model proposed by~\cite{sun2021multimodal}. We also adapted ViLT~\cite{kim2021vilt}, a robust VLP model, to perform multi-modal dialog intent prediction.

\paragraph{\SmallImage{fig/hexagon}~\textbf{Multi-Modal Dialog Retrieval~(T2I)}}
SCAN~\cite{lee2018stacked} is a cross-attention model that captures interactions between image regions and text tokens for inferring image-text similarity. DE++~\cite{feng2022mmdialog,zang2021photochat} applies CLIP\cite{radford2021learning} encoders for text and image, with a ranking module for scoring candidate relevance.

\paragraph{\SmallImage{fig/circle}~\textbf{Multi-Modal Dialog Retrieval(I2T)}}
TransResNet~\cite{shuster2018image} adopts three sub-networks to construct its retrieval model, which model three modalities of inputs, including vision, dialogue history, and style. UTC~\cite{chen2022utc} introduces a unified transformer with inter-task contrastive Learning for visual Dialog.
\paragraph{\SmallImage{fig/square}~\textbf{Multi-Modal Dialog State Tracking}}
BART~\cite{lewis2019bart} uses a seq2seq model with a bidirectional encoder and a left-to-right decoder.  ~\citet{liao2021mmconv} adopt DS-DST~\cite{zhang2019find} from textual DST and make use of image information via predicted labels. 
\paragraph{\SmallImage{fig/triangle}~\textbf{Multi-Modal Response Generation}}
SimpleTOD~\cite{hosseini2020simple} is a simple approach to task-oriented dialogue that approaches all of task-oriented dialogue as a single sequence generation problem. SimpleTOD can then directly leverage pre-trained models like GPT-2 to transfer language understanding from open-domain settings where data is more readily available. 

\section{Details For VDscore Metric}
\label{detail_vdscore}
Following the AHP method, first, we calculate the corresponding weight coefficients of each task.
\setlength{\arraycolsep}{2.3pt}
\begin{spacing}{1.0} 
\begin{gather*}
    \begin{bmatrix}
        1 & 2 & 3 & 3 & 5 \\
        1/2 & 1 & 2 & 2 & 4 \\
        1/3 & 1/2 & 1 & 1 & 3 \\
        1/3 & 1/2 & 1 & 1 & 3\\
        1/5 & 1/5 & 1/3 & 1/3 & 1
    \end{bmatrix}
    \xrightarrow[\text{by col}]{\text{normalized}}
    \begin{bmatrix}
        0.42 & 0.47 & 0.41 & 0.41 & 0.31 \\
        0.21 & 0.24 & 0.27 & 0.27 & 0.25 \\
        0.14 & 0.12 & 0.14 & 0.14 & 0.19 \\
        0.14 & 0.12 & 0.14 & 0.14 & 0.19 \\
        0.09 & 0.06 & 0.05 & 0.05 & 0.06 \\
    \end{bmatrix}
    \xrightarrow[\text{by row}]{\text{summed}}
    \begin{bmatrix}
        2.02 \\
        1.24 \\
        0.72 \\
        0.72 \\
        0.30 \\
    \end{bmatrix}
    \xrightarrow[\text{by col}]{\text{normalized}}
    \begin{bmatrix}
        0.41 \\
        0.25 \\
        0.14 \\
        0.14 \\
        0.06  \\
    \end{bmatrix} \\
\end{gather*}
\end{spacing}
Then, we separately calculate $\lambda$, $n$ and CI, where $\lambda$ represents the maximum eigenvalue of the pairwise comparison matrix, $n$ represents the number of all unique non-zero eigenvalues, and the consistency index CI is calculated from a and n as a consistency measure. 
\begin{spacing}{1.0}
\begin{gather*}
    \begin{bmatrix}
        1 & 2 & 3 & 3 & 5 \\
        1/2 & 1 & 2 & 2 & 4 \\
        1/3 & 1/2 & 1 & 1 & 3 \\
        1/3 & 1/2 & 1 & 1 & 3\\
        1/5 & 1/5 & 1/3 & 1/3 & 1
    \end{bmatrix}
    \times
    \begin{bmatrix}
        0.41 \\
        0.25 \\
        0.14 \\
        0.14 \\
        0.06  \\
    \end{bmatrix}
    =
    \begin{bmatrix}
    2.06 \\
    1.26 \\
    0.73 \\
    0.73 \\
    0.30 \\
    \end{bmatrix}
    \begin{array}{c}
        \lambda=\frac{1}{5} \times (\frac{2.06}{0.41}+\frac{1.26}{0.25}+\frac{0.73}{0.14}+\frac{0.73}{0.14}+\frac{0.30}{0.06})=5.07 \vspace{1ex} \\ 
        n = 5\\
        \ \\
        CI=\frac{\lambda-n}{n-1}=\frac{5.07-5}{5-1}=0.014 \vspace{1ex} \\
    \end{array} \\
\end{gather*}
\end{spacing}
Finally, we calculate the consistency ratio CR, and RI value referenced Table~\ref{tab:RI}:
\begin{equation}
  CR=\frac{CI}{RI(n=5)}=\frac{0.014}{1.12}=0.013\quad  \\  
\end{equation}
Because the consistency ratio (CR) is less than 0.1, it is considered to have passed the consistency check.
\begin{table}[b]
\centering
  \resizebox{0.60\textwidth}{!}{
  \begin{tabular}{cccccccccccc}
    \toprule
    \textbf{n}  & \textbf{1} & \textbf{2} & \textbf{3} & \textbf{4} & \textbf{5} & \textbf{6} & \textbf{7} & \textbf{8} & \textbf{9} & \textbf{10} & \textbf{11}\\
    \midrule
     \textbf{RI} & 0.00 & 0.00 & 0.58 & 0.90 & 1.12 & 1.24 & 1.32 & 1.41 & 1.45 & 1.49 &1.51 \\
  \bottomrule
\end{tabular}}
\caption{Reference table for Randomness Index (RI) metrics.}
\label{tab:RI}
\end{table}

\end{document}